\documentclass{article}
\usepackage{ijcai22}

\usepackage{times}
\usepackage{soul}
\usepackage{url}
\usepackage[hidelinks]{hyperref}
\usepackage[utf8]{inputenc}
\usepackage[small]{caption}
\usepackage{graphicx}
\usepackage{amsmath}
\usepackage{amsthm}
\usepackage{booktabs}
\usepackage{algorithm}
\usepackage{algorithmic}
\usepackage{multirow}
\usepackage{subcaption}
\usepackage{amsfonts}
\usepackage[table,xcdraw]{xcolor}

\newcommand{\xu}[1]{\textcolor{black}{#1}}

\title{TerViT: An Efficient Ternary Vision Transformer}

\author{
Sheng Xu$^1$\and
Yanjing Li$^1$\and
Teli Ma$^{2}$\and
Bohan Zeng$^1$\and
Baochang Zhang$^1$\and
Peng Gao$^2$\and
Jinhu L{\"u}$^1$
\affiliations
$^1$Beihang University\\
$^2$Shanghai Artificial Intelligence Laboratory
}

\begin{document}

\maketitle

\begin{abstract}
Vision transformers (ViTs) have demonstrated great potential in various visual tasks, but suffer from  expensive computational and memory cost problems when  deployed on resource-constrained devices. \xu{In this paper, we introduce a ternary vision transformer (TerViT) to ternarize the weights in ViTs,} which are challenged  by the large loss surface gap between real-valued and ternary parameters. To address the issue, we introduce a progressive training scheme by first training 8-bit transformers and then TerViT, and achieve a better optimization than conventional methods.  Furthermore, we introduce channel-wise ternarization, by partitioning each matrix to different channels, each of which is with an unique distribution and ternarization interval. We apply our methods to popular DeiT and Swin backbones, and extensive results show that we can achieve competitive performance. For example, TerViT can quantize Swin-S to  13.1MB model size while achieving above 79\% Top-1 accuracy on ImageNet dataset. 
\end{abstract}

\section{Introduction}

Inspired by the success in Natural Language Processing (NLP) tasks, transformer-based models have shown great power in various Computer Vision (CV) tasks, such as image classification \cite{vit} and object detection \cite{detr}. Pre-trained with large-scale data, these models usually have tremendous number of parameters. For example, there are 632M parameters taking up 2528MB memory usage and 162G FLOPs in the ViT-H model, which is both memory and computation expensive during inference. This limits these models for the deployment on resource-limited platforms. Therefore,   compressed transformers are urgently needed for real applications.

Substantial efforts have been made to compress and accelerate neural networks for efficient online inference. Methods include compact network design \cite{mobilenet}, network pruning \cite{he2018soft}, low-rank decomposition \cite{denil2013predicting}, quantization \cite{qin2020forward}, and knowledge distillation \cite{romero2014fitnets}. Quantization is particularly suitable for deployment on AI chips because it reduces the bit-width of network parameters and activations for efficient inference. Prior post-training quantization (PTQ) methods \cite{liu2021post} on ViTs directly compute quantized parameters based on pre-trained full-precision models, which constrains model performance to a sub-optimized level without fine-tuning. Furthermore, quantizing these models  based on PTQ methods to ultra-low bits ({\em e.g.}, 1 or 2 bits) is inefficient and suffer from significant performance reduction.
\begin{figure}
	\centering
	\includegraphics[scale=.8]{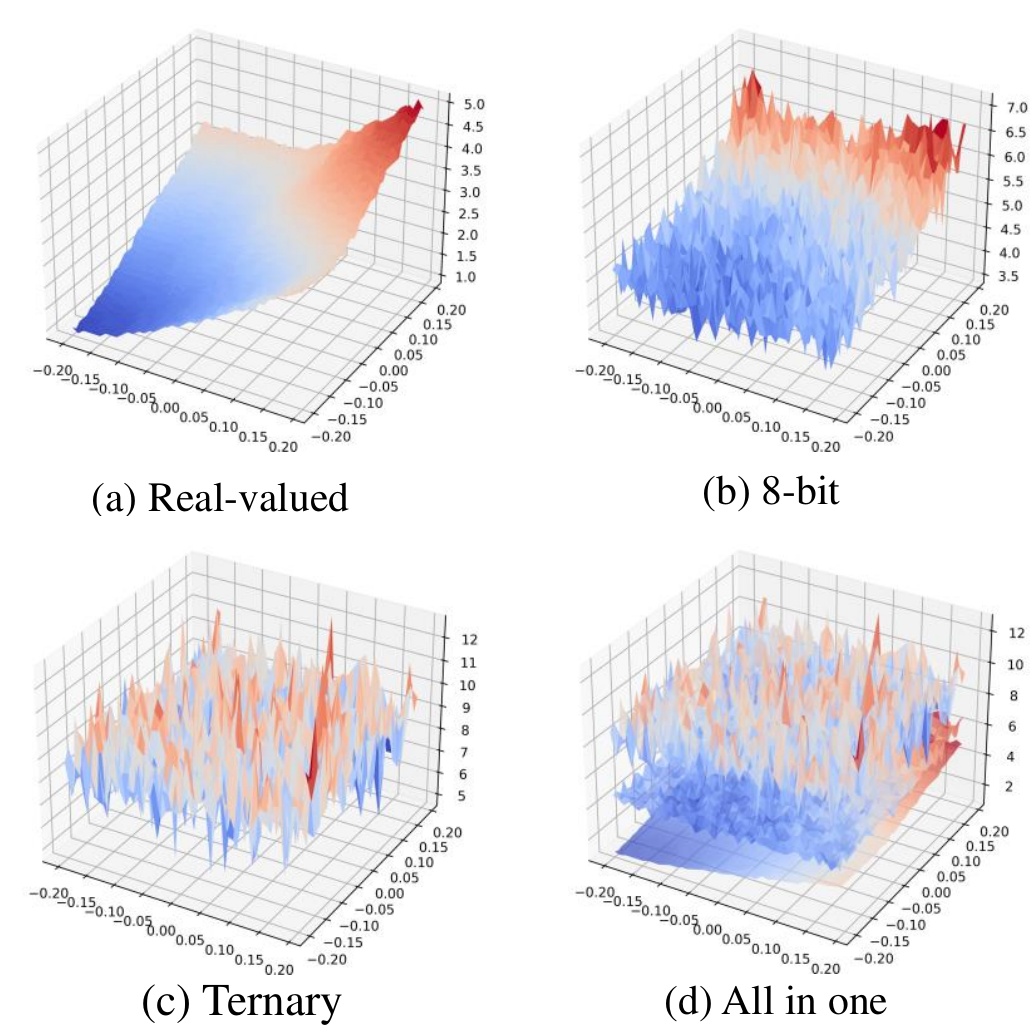}\\
	\caption{The actual optimization landscape from (a) real-valued, (b) 8-bit, and (c) ternary DeiT-S backbone. (d) shows the gap among the three surfaces by stacking them together.} 
	
	\label{motivation}
\end{figure}
\begin{figure*}[t]
	\centering
	\includegraphics[scale=.45]{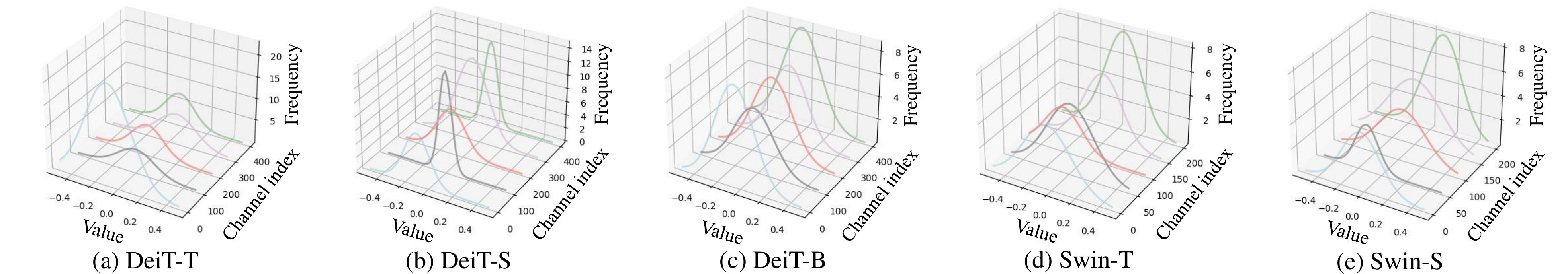}\\
	\caption{Channel-wise weight distribution of the first MHSA layer of (a) DeiT-T, (b) DeiT-S, (c) DeiT-B, (d) Swin-T, and (e) Swin-S. We select the 1-st, 100-th, 200-th, 300-th and 400-th channels to visulize. It is obviously shown that the channel-wise distribution variance varies to a large extent, which motivates us to introduce channel-wise ternarization.} 
	\label{dist}
\end{figure*}

Differently, quantization-aware training (QAT) \cite{liu2020reactnet} methods perform quantization during back propagation and achieve much lower performance drop and generally higher compression rate. QAT is shown to be effective for 
CNN models \cite{liu2018bi} for CV tasks. However,  QAT methods remain unexplored for ternary quantization of vision transformers, \xu{due to the global information extracting mechanism and densely connected structure. }

In this paper, we following the visualizing method in \cite{li2018visualizing}, and observe that the commonly used QAT methods \cite{shen2020q,zhang2020ternarybert} are, such as fine-tuning ternary vision transformers from real-valued weights,  ineffective for vision transformers. In Fig. \ref{motivation}(a),  the real-valued model  has the lowest overall training loss, whose  loss landscape is flat and robust to the perturbation. For the 8-bit model (Fig. \ref{motivation}(b)), despite the surface tilts up with larger perturbations, it looks locally convex and this may  explain why the ViT model can be 8-bit quantized without any fine-tuning and severe accuracy drop \cite{liu2021post}. However, the loss landscape of the ternary model (Fig. \ref{motivation}(c)) is more complex with numerous local minimas.  By stacking  three landscapes together (Fig. \ref{motivation}(d)), the loss surface of the ternary model stands on the top with a large margin against the other two. The steep curvature of loss surface reflects a higher sensitivity to binarization, which can attribute to the training difficulty. However, this phenomena inspires us that the 8-bit model can be utilized to bridge the optimization gap between the real-valued model and the ternary model.







Based on the above analysis, we introduce an efficient Ternary Vision Transformer (TerViT), aiming at improving the loss landscape of TerViTs for a better optimization. We progressively take the 8-bit model as a proxy to bridge the gap between the ternary and real-valued models. Specifically, ternary weights converts the latent full-precision weights into a proxy 8-bit model to initialize the training of TerViT. Moreover, we introduce channel-wise ternarization, considering that the different distribution range of real-valued weights from channel to channel.
In a summary, we address the following issues and our contributions are as 
\begin{itemize}
    \item We introduce a ternary vision transformer for the quantization of vision transformer for the first time.  We provide a deep investigation into our progressive training method by introducing multiple metrics to analyze the convergence of the optimization.
    \item We introduce a channel-wise quantization scheme, which can improve the quantization stability without  increasing the model complexity. 
    \item \xu{Experiment results show that our method achieves the  state-of-the-art quantized vision transformer. For example, on the  Swin-S backbone, we achieve 15.25$\times$ compression ratio in terms of model size, 4$\times$  smaller activation size, within 4\% accuracy loss.} 
\end{itemize}
\vspace{-3mm}
\section{Methodology}
This section describes our TerViT \xu{in details}. We first overview  vision transformers \xu{and then} describe our quantization framework, including channel-wise weight ternarization and 8-bit activation quantization. 
We \xu{elaborate} the 'dead weight' problem and visualize the two-dimensional loss landscapes to analyze how our progressive training helps conquer the zero-gradient local minima.

\subsection{Preliminary}
\begin{figure*}[t]
	\centering
	\hspace{-3mm}
	\includegraphics[scale=.55]{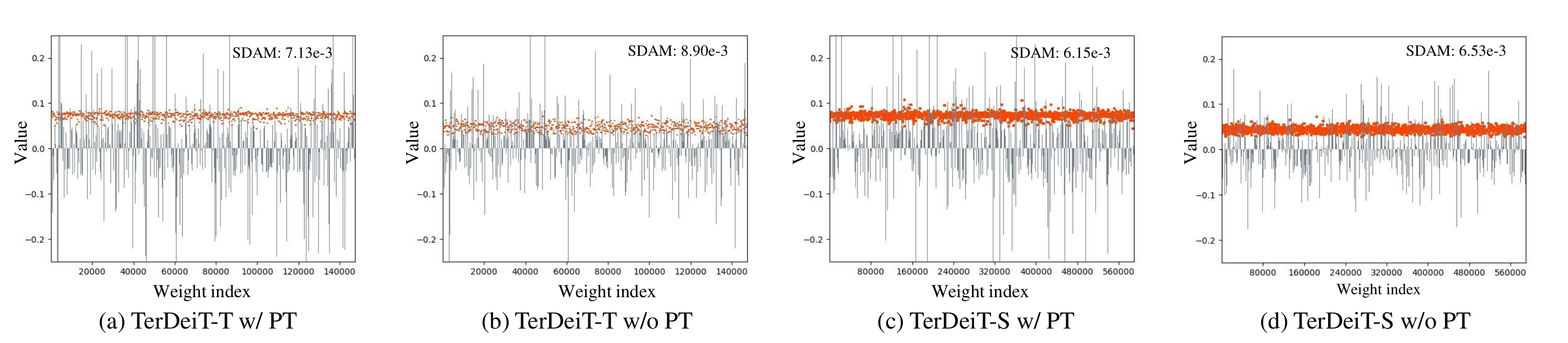}\\
	\caption{The latent weight value distribution in the first ternary fully-connected layer of MLP after training one epoch. For clarity, we use red hyphens to mark the Channel-wise Absolute Mean (CAM) of real-valued weights in each channel. Compared to simple fine-tuning, our progressive training (PT) boosts the ternary ViTs to much lower CAM value and higher Standard Deviation (SDAM), which indicates that the model model can be trained better with PT than those who are directly fine-tuned with real-valued model.} 
	\label{cam}
\end{figure*}
\begin{figure*}
	\centering
	\hspace{-3mm}
	\includegraphics[scale=.55]{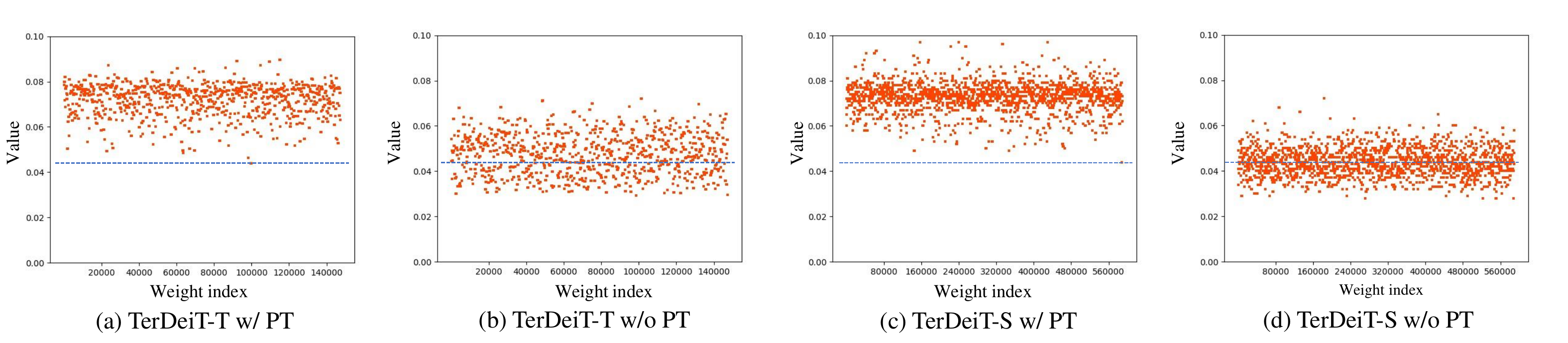}\\
	\caption{The update value distribution of weights in the first ternary fully-connected layer of MLP after trained with one epoch. For clarity, we omit the original update value distribution and use red hyphens to mark the CAM of the weights’ update values in each channel. The grey dotted line denotes the minimum CAM value of weights in the TerViT with PT. In this layer, 36.7\% (DeiT-T) and 53.7\% (DeiT-S) of the channels  with real-valued pre-trained have a lower CAM than the minimum CAM with PT.} 
	\label{cam2}
\end{figure*}
A standard transformer block includes two main modules: Self Attention (SA) module and
Multi-Layer Perceptron (MLP) module. For  a specific transformer layer, supposed its input is denoted as $f_{in}\in\mathbb{R}^{n\times d}$, the corresponding query, key, and value are calculated as
\begin{equation}
	\begin{aligned}
	f_{Q} = f_{in}W_Q, \quad
	f_{K} = f_{in}W_K, \quad
	f_{V} = f_{in}W_V,
	\end{aligned}	
	\label{qkv}
	\end{equation}
where $W_Q$, $W_K$, $W_V$ denote matrixes to generate queries, keys and values respectively. Then, the attention scores computed by the dot product of queries and keys can be formulated as
\begin{equation}
	{\bf A} = \operatorname{softmax}(\frac{f_{Q}f_{K}^T}{\sqrt{d}}).
	\label{attn}
	\end{equation}
Finally, we calculate the weighted sum of attention weights $A$ and $f_V$, thus obtaining integrated features as
\begin{equation}
	\label{self-attention}
	\begin{aligned}
	f_{out} &= {\bf A} f_V\cdot W_O, 
	\end{aligned}
\end{equation}
where $W_O$ denotes the projection matrix. The MLP module contains two linear layers parameterized by $W_1 \in \mathbb{R}^{d\times (\mathcal{E} d)}$, $b_1\in\mathbb{R}^{\mathcal{E} d}$ and $W_2\in\mathbb{R}^{(\mathcal{E} d)\times d}$, $b_2\in\mathbb{R}^{d}$ respectively, where $\mathcal{E}$ is the expand ratio of MLP layers. Denote the input to MLP as $f_{in}\in\mathbb{R}^{n\times d}$, the output is then computed as
\begin{equation}
	f_{out} = \operatorname{GeLU}({f_{in}W_1+b_1})W_2+b_2. 
	\label{out}
\end{equation}
The most computational costs of vision transformer lie in the large matrix multiplication in SA and MLP module. Following the mainstream quantization methods for CNNs \cite{rastegari2016xnor}, we quantize all the weights and inputs involved in matrix multiplication. For weight, we ternarize the weights $W_Q, W_K, W_V, W_O, W_1, W_2$ in Equ. \ref{qkv} and \ref{out} for all transformer layers. Besides these weights, we also quantize the inputs of all linear layers and matrix multiplication operations into 8-bit. Following the methods in \cite{zhang2020ternarybert}, we do not quantize the softmax operation and layer normalization, because the parameters contained in these operations are negligible and quantizing them may bring significant accuracy degradation.

\begin{figure*}
	\centering
	\hspace{-3mm}
	\includegraphics[scale=.43]{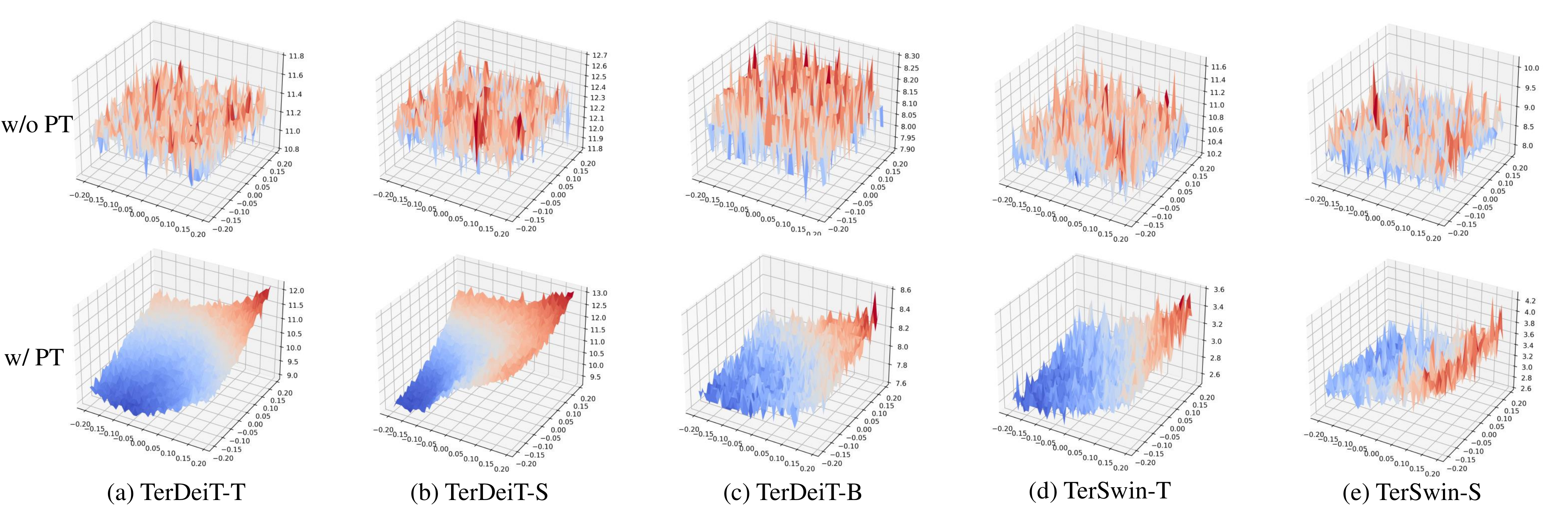}\\
	\caption{Comparison of the loss landscapes of all models we employed between w/o PT and w/ PT. The situation w/o PT conquers more zero gradient local minima.} 
	\label{optim}
	\vspace{-4mm}
\end{figure*}
\subsection{Quantization \xu{for Vision Transformers}}
\xu{In this section, we address key issues in quantization for ViTs, including a channel-wise weight ternarization together with a comprehensive investigation into weights distribution range, and also the activation quantization method in details.}

\noindent{\bf Channel-wise weight ternarization.} Following the Equs. \ref{qkv} $\sim$ \ref{out}, the input feature map has $n$ patches and each patch has $d$-dim embedding channels. In the multi-head self-attention (MHSA) module, each self-attention head has 3 weight matrices, {\em i.e.}, $W_Q, W_K, W_V \in \mathbb{R}^{\frac{d}{N_h}\times d}$, where $N_h$ is the number of attention heads. In the MHSA module, the computation process is rewritten as 

\begin{equation}
    {\bf A}^i = \operatorname{softmax}(\frac{f_{in}W^i_Q{W^i_K}^T{f_{in}}^T}{\sqrt{d}}),
\end{equation}
and
\begin{equation}
    f_{out} = \operatorname{concatenate}\{{\bf A}^1 W_V^1, \cdots, {\bf A}^{N_h} W_V^{N_h}\}\cdot W_O.
\end{equation}
Directly quantizing each 3 matrices in MHSA as an entirety with the same quantization range can significantly degrade the accuracy, since there are $3\times n\times d$ parameters in total, and the weights corresponding to each channel may lie in different range of real-valued numbers.
	
As shown in Fig. \ref{dist}, the distribution range of real-valued weights in a MHSA module  \xu{vary} from channel to channel with non negligible differences. Thus, in \xu{previous work \cite{li2016ternary}}, the layer-wise ternarization quantizes all parameters in a single layer with the same threshold $\Delta_W$ dismissing the channel-wise feature representation ability of  ViTs, leading to performance degradation. Based on the discussion and observation mentioned above, we introduce a channel-wise ternarization as
\begin{equation}
{\bf T}^{W} = \alpha \circ \operatorname{Ternarize}(W),
\label{t}
\end{equation}
where $\circ$ denotes the channel-wise multiplication. $\alpha$ is the channel-wise scale factor defined by the channel-wise absolute mean (CAM) as 
\begin{equation}
    \alpha^j = \frac{1}{n}\sum^n_{k=1}|W^{k,j}|,
\end{equation}
and the ternarization function $\operatorname{Ternarize}(\cdot)$ \xu{in Equ. \ref{t}} is further defined as 
\begin{equation}
\small
\label{ternary}
\begin{aligned}
	&\operatorname{Ternarize}(W^j) = \left\{
	\begin{aligned}
	& -1 & W^j < -\Delta_{W^j} \\
	& 0 & - \Delta_{W^j} \leq W^j < \Delta_{W^j}\\
	& +1  & W^j \geq \Delta_{W^j}\\
	\end{aligned}
	\right. 
	\end{aligned}
	\end{equation}
where the real-valued weights are ternarized with a threshold $\Delta_{W^j} = 0.7 \frac{\|W^j\|_{1}}{n}$ corresponding to the distribution range of real-valued parameters. 


\noindent{\bf Activation quantization.} To make the most expensive matrix multiplication operation faster, following the prior works of NLP task \cite{shen2020q}, we also quantize  activations, {\em i.e.}, inputs of all linear layers ($f_{in}$) and matrix multiplication ($f_Q$, $f_K$, $f_V$ and ${\bf A}$) to 8 bits. There are two kinds of commonly used 8-bit quantization methods: symmetric and min-max 8-bit quantization. The quantized values of the symmetric 8-bit quantization distribute symmetrically in both sides of 0, while those of min-max 8-bit quantization distribute uniformly in a range determined by the minimum and maximum values. Specifically, for one element $x$ in the activation $f$, denote $x_{max} = max(f)$ and $x_{min} = min(f)$, the min-max 8-bit quantization function is
\begin{equation}
    \label{activation}
	\begin{aligned}
	&{\bf Q}^{f} = round(\frac{f-x_{min}}{s})\times s + x_{min},\\
	\end{aligned}
	\end{equation}
where $s=\frac{x_{max}-x_{min}}{255}$ is the quantization scale.

\subsection{Progressive Training}
In this \xu{section}, we illustrate the phenomenons and statistics to show the benefit of our proposed progressive training (PT) on vision transformers.

\noindent\textbf{Reducing dead weights.} 'Dead weights' is the phenomenon that the weights in some channels are not optimized to learn meaningful representations, which are intuitively described as 'dead'. In these channels, the gradient will always stay small \cite{xu2021recu,liu2021adam}, which causes insufficient training. Note that the weights refer to the real-valued latent weights, {\em i.e.}, $W_Q$, $W_K$, $W_V$, $W_O$, $W_1$, and $W_2$. The magnitude of these real-valued weights are regarded as ‘inertial’ \cite{helwegen2019latent}, indicating how likely the corresponding ternary weights are going to change the value. 

An universal measurement of the 'dead weights' is the Channel-wise Absolute Mean (CAM), which captures the average amplitude of real-valued weights within a kernel. The value of CAM is the same channel-wise scale factor as defined in Sec. 3.2. As shown in Fig. \ref{cam}, we observe that the CAM of latent weights in the same networks without PT are small in their values compared with the one with PT. Another quantitative result is shown in Fig. \ref{cam2}, the DeiT-T and DeiT-S without PT generates 36.7\% and 53.7\% CAM lower than the lowest CAM of the counterparts with PT, respectively. Thus, there exists unbalanced weight fine-tuning when the model is trained without \xu{PT}. 

To further measure the distribution of the trained latent real-valued weight magnitude, the Standard Deviation of the Absolute Mean (SDAM) of the real-valued weight magnitude on each output channel is also calculated in Fig. \ref{cam}. It is evident that the SDAM of fine-tuning with PT is lower than that of fine-tuning without PT, revealing much fewer 'dead weights' when fine-tuning the model with PT.

\noindent\textbf{Optimization.} For \xu{a better illustration}, we plot a two-dimensional loss surface of the networks with two training methods which are distinguished by whether PT method is employed, following \cite{li2018visualizing}. As shown in the top line in Fig. \ref{optim}, directly ternarizing real-valued pretrained models is challenging, for the cluttered and non-convex loss landscape. To facilitate the optimization of TerViT, our PT method first quantized the pre-trained real-valued models to 8-bit for reducing the performance gap between full-precision weights and low-precision counterparts. After training 8-bit models for specific epochs, the parameters can be smoothly transferred to 2-bit without tremendous performance drop and loss increase and achieve more stable quantized process. According to the bottom line in Fig. \ref{optim}, our PT method can optimize the ternary models to obtain better results and be more robust with loss landscape closer to convex function, compared to  baseline training methods. 

\section{Experiments}
In this section, we evaluate the performance of the proposed TerViT model for image classification task using popular DeiT \cite{touvron2021training} and Swin \cite{liu2021swin} backbones. To the best of our knowledge, there is no published work done on quantization-aware training of vision transformer at this point, so we implement the baseline TWN \cite{li2016ternary} methods for CNNs as described in the papers by ourselves. It is shown that the proposed method outperforms the conventional TWN method and even achieves comparable performance with significant compression ratio as the post-training quantization methods on some models. Moreover, extensive experiments of ablation study have shown that the proposed progressive training and channel-wise quantization method are beneficial for the ternary vision transformer.

\begin{table}[t]
\centering
\caption{Evaluation of the setups of progressive training (PT). 'From scratch' denotes directly training ternary networks from scratch for 300 epochs. 'From Real-valued' denotes training ternary networks from real-valued pre-trained models for 300 epochs, {\em i.e.}, without PT. '(a, b)' denotes the step size of 8-bit pre-training and ternary fine-tuning, respectively. }
\footnotesize
\setlength{\tabcolsep}{0.1mm}{\begin{tabular}{c|c|c|cccc}
\hline
\multirow{2}{*}{Model} & \multirow{2}{*}{\begin{tabular}[c]{@{}c@{}}From\\ scratch\end{tabular}} & \multirow{2}{*}{\begin{tabular}[c]{@{}c@{}}From\\ real-valued\end{tabular}} & \multicolumn{4}{c}{Setup}                        \\ \cline{4-7} 
                       &                                                                         &                                                                             & (50,250)      & (100,200) & (150,150) & (200,100) \\ \hline
TerDeiT-T              & 59.3                                                                    & 65.0                                                                        & \textbf{66.6} & 66.1      & 65.3      & 64.6      \\
TerDeiT-S              & 66.9                                                                    & 72.1                                                                        & \textbf{74.2}          & 73.7      & 72.9      & 71.2      \\ \hline
\end{tabular}}
\label{setup}
\end{table}

\begin{table}[]
\centering
\caption{The effects of different components in TerViT using DeiT-T backbone on the final accuracy. We select TWN as the baseline method. 'PT' denotes the progressive training. The first and last layer are quantized to 8-bit.}
\begin{tabular}{cc}
\hline
Method                       & Top-1 Accuracy \\ \hline
Real-valued                  & 72.2           \\ \hline
TWN (layer-wise)             & 64.4           \\
TWN + channel-wise             & 65.0           \\
TWN + PT                       & 65.8           \\
TWN + channel-wise + PT (TerViT) & \textbf{66.6}           \\ \hline
\end{tabular}
\label{components}
\vspace{-3mm}
\end{table}

\subsection{Datasets and Implementation Details}
\noindent{\bf Datasets.} The experiments are carried out on the ILSVRC12 ImageNet classification dataset \cite{imagenet12}, which is more challenging than small datasets such as CIFAR \cite{krizhevsky2009learning} and MNIST \cite{netzer2011reading}. The ImageNet dataset is more challenging due to its large scale and greater diversity. There are 1000 classes and 1.2 million training images, and 50k validation images in it. In our experiments, we use the classic data augmentation method described in \cite{touvron2021training}.

\noindent{\bf Experimental settings.} Given a well-trained real-valued ViT model, we first fine-tune it with 8-bit quantization on both weights and activations for 50 epochs. Then, we convert the 8-bit weight quantization to ternarization by 250 epochs. Thus, the PT method fine-tunes the models for total 300 epochs. For fair comparison, we directly fine-tune the counterparts without PT 300 epochs to validate effectiveness of our PT. The original TWN \cite{li2016ternary} method is employed to validate the effectiveness of our channel-wise quantization. We quantize the first layer (patch embedding) and last layer (classification head) to 8 bits. The training hyper-parameters is selected following \cite{touvron2021training}. The AdamW optimizer is employed. 

\noindent{\bf Baseline.} We evaluate our quantization method on two popular vision transformer implementation: DeiT \cite{touvron2021training} and Swin \cite{liu2021swin}. The DeiT-T, DeiT-S, DeiT-B, Swin-T and Swin-S are adopted as the baseline model, whose Top-1 accuracy on ImageNet dataset are 72.2\%, 79.9\%, 81.8\%, 81.2\%, and 83.2\% respectively. For a fair comparison, we utilize the official implementation of DeiT and Swin, without using other techniques like knowledge distillation.  

\subsection{Ablation Study}
\noindent{\bf Selecting progressive training setups.} 
Progressive training (PT) is the main contribution of our paper, we evaluate the different setup of PT, {\em i.e.}, the length of the 8-bit pre-training and ternary network fine-tuning. As shown in the first 3 columns of Tab. \ref{setup}, fine-tune from pre-trained model boosts the performance by about 5\% $\sim$ 6\% Top-1 accuracy. As shown in the last columns, the best performance occurs when we first fine-tune the 8-bit model for 50 epochs and then train the ternary model for 250 epochs.

\noindent{\bf Evaluating the components.} 
In this part, we evaluate every critical part of TerViT to show how we compose the novel and effective TerViT. We first introduce our baseline network, {\em i.e.}, TWN \cite{li2016ternary}, achieving 64.4\% Top-1 accuracy. As shown in Tab. \ref{components}, the introduction of channel-wise ternarization and PT improves the accuracy by 0.6\% and 1.4\% respectively over the  baseline network, as shown in the second section of Tab. \ref{components}. By adding all the channel-wise ternarization and PT, our TerViT achieves 2.2\% higher accuracy than the baseline, notably narrowing the gap between ternary DeiT-T and the real-valued counterpart. 

\begin{table}[t]
\caption{Quantitative results of different bit-width of the patch embedding and head layer for TerDeiT-T model. We vary the bit-width of the two layers from 32-bit (real-valued) to 2-bit (ternary). The relevant model size and Top-1 accuracy are reported for comparison.} 
\centering
\setlength{\tabcolsep}{2mm}{\begin{tabular}{c|cccc}
\hline
Model                   & \begin{tabular}[c]{@{}c@{}}Patch\\ Embedding\end{tabular} & Head   & \begin{tabular}[c]{@{}c@{}}Model\\size (MB)\end{tabular} & \begin{tabular}[c]{@{}c@{}}Top-1\\Accuracy\end{tabular}\\ \hline
\multirow{5}{*}{TerDeiT-T} & 32-bit                                                    & 32-bit & 2.7             & 66.9  \\
                        & 8-bit                                                     & 32-bit & 2.2             & 66.7  \\
                        & 32-bit                                                    & 8-bit  & 2.1             & 66.7  \\
                        & \textbf{8-bit}                                                     & \textbf{8-bit}  & \textbf{1.6}             & \textbf{66.6}  \\
                        & 2-bit                                                     & 2-bit  & 1.4             & 44.2  \\ \hline
\end{tabular}}
\label{quant}
\end{table}
\begin{figure}[t]
	\centering
	\includegraphics[scale=.5]{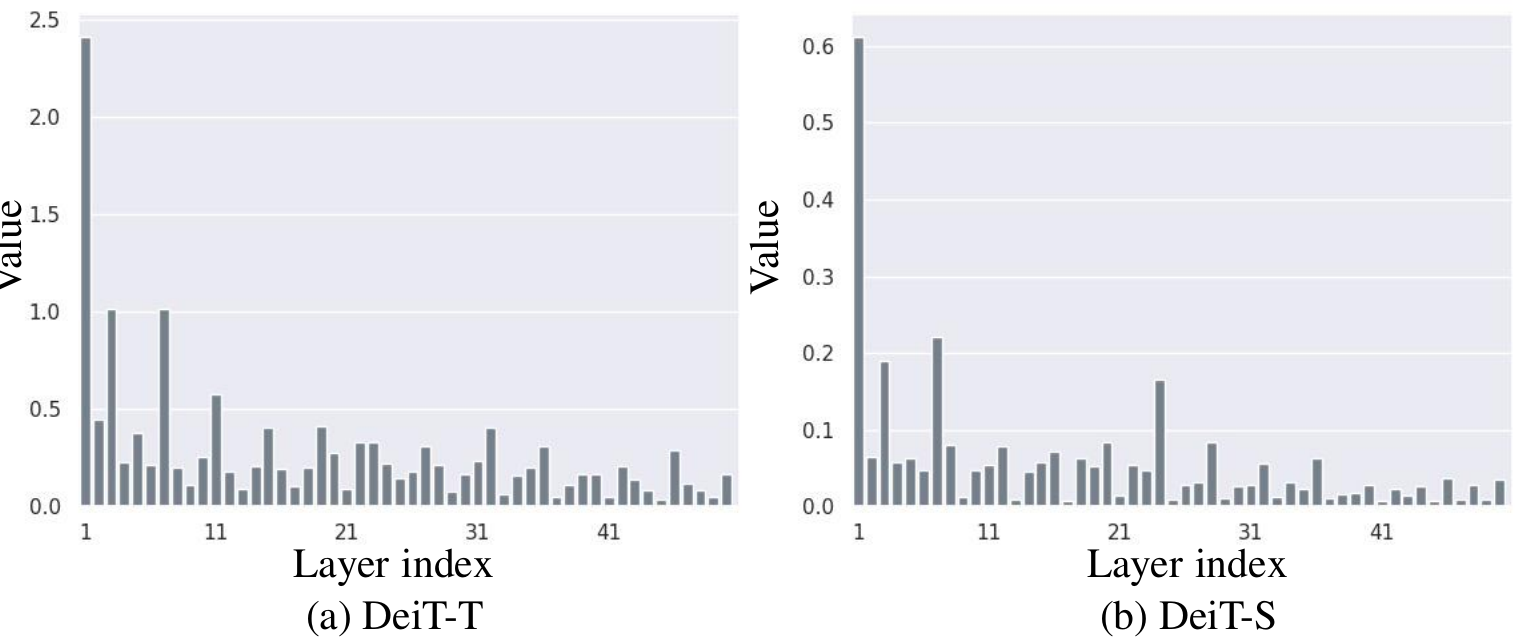}\\
	\caption{Top eigenvalue of each layer's Hessian matrix of pre-trained (a) DeiT-T and (b) DeiT-S on ImageNet dataset. The eigenvalue of the patch embedding layer is obviously larger than the other layers, indicating more sensitivity.}
	\label{hessian}
	\vspace{-3mm}
\end{figure}
\begin{table*}[t]
\small
\centering
\caption{Quantization results on ImageNet dataset. We abbreviate quantization bits used for weights as ‘W-bits’ and activations as ‘A-bits’. In particular, we first compare with the 8-bit approaches. Then we compare ours with the 2-bit baseline TWN. Our method achieves acceptable testing performance drop with significantly high compression ratio. Also note that we use 8-bit for the first and last layers.}
\begin{tabular}{c|cccccc}
\hline
Model                    & Method                         & W-bit                     & A-bit                     & Model size (MB)              & Compression Ratio                     & Top-1 Accuracy               \\ \hline
                         & Real-valued                     & 32                        & 32                        & 22.7                         & -                                     & 72.2                         \\ \cline{2-7} 
                         & MinMax                         & 8                         & 8                         & 5.7                          & 3.98$\times$                          & 67.9                         \\
                         & Percentile                     & 8                         & 8                         & 5.7                          & 3.98$\times$                          & 69.5                         \\ \cline{2-7} 
                         & TWN                            & 2                         & 8                         & 1.6                          & 13.35$\times$                         &     64.4                         \\
\multirow{-5}{*}{DeiT-T} & \cellcolor[HTML]{C0C0C0}TerViT & \cellcolor[HTML]{C0C0C0}2 & \cellcolor[HTML]{C0C0C0}8 & \cellcolor[HTML]{C0C0C0}1.6  & \cellcolor[HTML]{C0C0C0}13.35$\times$ & \cellcolor[HTML]{C0C0C0}66.6\color{blue}{(-5.6)} \\ \hline
                         & Real-valued                    & 32                        & 32                        & 88.2                         & -                                     & 79.9                         \\ \cline{2-7} 
                         & Percentile                     & 8                         & 8                         & 22.2                         & 3.97$\times$                          & 74.0                         \\
                         & VT-PTQ                         & 8                         & 8                         & 22.2                         & 3.97$\times$                          & 77.5                         \\ \cline{2-7} 
                         & TWN                            & 2                         & 8                         & 6.0                          & 14.70$\times$                         &            70.2                  \\
\multirow{-5}{*}{DeiT-S} & \cellcolor[HTML]{C0C0C0}TerViT & \cellcolor[HTML]{C0C0C0}2 & \cellcolor[HTML]{C0C0C0}8 & \cellcolor[HTML]{C0C0C0}6.0  & \cellcolor[HTML]{C0C0C0}14.70$\times$ & \cellcolor[HTML]{C0C0C0}74.2\color{blue}{(-5.7)} \\ \hline
                         & Real-valued                    & 32                        & 32                        & 346.2                        & -                                     & 81.9                         \\ \cline{2-7} 
                         & Percentile                     & 8                         & 8                         & 90.6                         & 3.82$\times$                          & 77.5                         \\
                         & VT-PTQ                         & 8                         & 8                         & 90.6                         & 3.82$\times$                          & 81.3                         \\ \cline{2-7} 
                         & TWN                            & 2                         & 8                         & 22.7                         & 15.25$\times$                         &           72.9                   \\
\multirow{-5}{*}{DeiT-B} & \cellcolor[HTML]{C0C0C0}TerViT & \cellcolor[HTML]{C0C0C0}2 & \cellcolor[HTML]{C0C0C0}8 & \cellcolor[HTML]{C0C0C0}22.7 & \cellcolor[HTML]{C0C0C0}15.25$\times$ &  \cellcolor[HTML]{C0C0C0}76.1\color{blue}{(-5.8)}  \\ \hline
                         & Real-valued                    & 32                        & 32                        & 114.2                        & -                                     & 81.4                         \\ \cline{2-7} 
                         & Percentile                     & 8                         & 8                         & 28.5                         & 3.99$\times$                           & 78.8                         \\
                         & OMSE                           & 8                         & 8                         & 28.5                         & 3.99$\times$                           & 79.3                         \\ \cline{2-7} 
                         & TWN                            & 2                         & 8                         & 7.7                          & 14.83$\times$                         &    73.7                         \\  
\multirow{-5}{*}{Swin-T} & \cellcolor[HTML]{C0C0C0}TerViT & \cellcolor[HTML]{C0C0C0}2 & \cellcolor[HTML]{C0C0C0}8 & \cellcolor[HTML]{C0C0C0}7.7  & \cellcolor[HTML]{C0C0C0}14.83$\times$ & \cellcolor[HTML]{C0C0C0}77.5\color{blue}{(-3.9)}  \\ \hline
                         & Real-valued                    & 32                        & 32                        & 199.8                        & -                                     & 83.2                         \\ \cline{2-7} 
                         & Percentile                     & 8                         & 8                         & 49.9                         & 3.99$\times$                          & 79.2                         \\
                         & OMSE                           & 8                         & 8                         & 49.9                         & 3.99$\times$                          & 79.6                         \\ \cline{2-7} 
                         & TWN                            & 2                         & 8                         & 13.1                         & 15.25$\times$                         &       76.1                      \\
\multirow{-5}{*}{Swin-S} & \cellcolor[HTML]{C0C0C0}TerViT & \cellcolor[HTML]{C0C0C0}2 & \cellcolor[HTML]{C0C0C0}8 & \cellcolor[HTML]{C0C0C0}13.1 & \cellcolor[HTML]{C0C0C0}15.25$\times$ & \cellcolor[HTML]{C0C0C0}79.5\color{blue}{(-3.7)}   \\ \hline
\end{tabular}
\label{imagenet}
\end{table*}
\noindent{\bf Quantizing the first and last layer.} 
Prior low-bit works in CNN \cite{liu2020reactnet} always save the first convolution layer and last fully-connected layer as real-valued due to the sensitivity. However, these real-valued layers counts a large proportion of the total model size. In our TerViT, we analyze the sensitivity of the first convolution layer (patch embedding layer) and last fully-connected layer (classification head) following \cite{dong2019hawq}. As shown in the Fig. \ref{hessian}, the Hessian matrix eigenvalue of the patch embedding layer is obviously larger than the other layers, which indicates that  the patch embedding layer is more sensitive than other layers. The classification head directly influences the output of network, thus we save it to 8-bit.  

We also conduct controlled experiments on the first and last layers using TerDeiT-T. In Tab. \ref{quant}, the 8-bit quantized first and last layers barely bring additional performance drop (only 0.3\%$\downarrow$) than real-valued (32-bit), offering a better performance-efficiency trade-off. When we set these two layers as 2-bit, the performance drops dramatically. Hence, we set these two layers to 8-bit for the extended experiments.


\subsection{Main Results}
The experimental results are shown in Tab. \ref{imagenet}. We compare our method with 2-bit baseline TWN \cite{li2016ternary} on the same framework for the task of image classification on the ImageNet dataset. We also report the classification performance of the 8-bit quantized networks percentile \cite{Li_2019_CVPR}, OMSE \cite{choukroun2019low}, and VT-PTQ \cite{liu2021post}. 

We firstly evaluate the proposed method on DeiT-T, DeiT-S and DeiT-B model. For DeiT-T backbone, compared with 8-bit percentile-based method, our TerDeiT-T achieves a much larger compression ratio than 8-bit percentile. However, the performance gap is rather small (66.6\% {\em vs.} 69.5\%). And it is worth noting that the proposed 2-bit model significantly compresses the DeiT-T by 13.35$\times$. The proposed method boosts the  performance of TWN by 2.2\% with the same architecture and bit-width, which is significant on the ImageNet dataset. For larger DeiT-S, as shown in Tab. \ref{imagenet}, the performance of the proposed method outperforms the TWN method by 4.0\%$\uparrow$, a large margin. Compared with 8-bit methods, our method achieves significantly higher compression rate, and the performance gap is rather small. For DeiT-B, as shown in Tab. \ref{imagenet}, the Top-1 accuracy of TWN is 72.9\%. And the proposed scheme improves the performance of the ternary model by 3.2\%$\uparrow$, up to 76.1\%. It is worth mentioning that the TerDeiT-B out performs real-valued DeiT-T backbone by 3.9\% with the same model size, which also proves the significance of our method.

Also, our method generates convincing results on Swin-transformers. As shown in Tab. \ref{imagenet}, the performance of the proposed method outperforms the TWN method by 3.8\%$\uparrow$ and 3.4\%$\uparrow$, a large margin. Compared with 8-bit methods, our method achieves significantly higher compression rate, and comparable performance. Note that our method achieves a small performance gap within 4\% compared with the real-valued counterpart using Swin transformers, which demonstrates the significance of our TerViT.

\section{Conclusion}
\xu{In this paper, we introduce} ternary vision transformers to improve the quantized ViTs with higher compression ratio and competitive performance. The presented TerViT introduces channel-wise ternarization and progressive training method. Notably, we also provide comprehensive and in-depth investigation in quantizing ViTs. As a result, the performance gap between TerViTs and real-valued counterparts can be significantly reduced. Extensive experiments validate the superiority of TerViT in image classification task compared with multiple mainstream backbones. Future work will focus on binarizing and more effective training methods on quantized vision transformers.

\bibliographystyle{named}
\bibliography{ijcai22}

\end{document}